# ASCAPE: An open AI ecosystem to support the quality of life of cancer patients

Konstantinos Lampropoulos
*Department of Electrical & Computer Engineering*
*University of Patras*
Patras, Greece
klamprop@ece.upatras.gr

Thanos Kosmidis
*CareAcross Research*
*CareAcross LTD*
London, UK
thanos.kosmidis@careacross.com

Serge Autexier
*RD Cyber-Physical Systems*
*German Research Center for Artificial Intelligence (DFKI)*
Bremen, Germany
serge.autexier@dfki.de

Miloš Savić
*Division of Informatics, Faculty of Sciences*
*University of Novi Sad*
Novi Sad, Serbia
svc@dmi.uns.ac.rs

Manos Athanatos
*Institute of Computer Science Foundation for Research and Technology – Hellas*
Heraklion, Crete, Greece
athanat@ics.forth.gr

Miltiadis Kokkonidis
*Research & Innovation Development INTRASOFT International S.A.*
Luxembourg, Luxembourg
miltiadis.kokkonidis@intrasoft-intl.com

Tzortzia Koutsouri
*SPHYNX Research*
*SPHYNX Technology Solutions AG*
Zug, Switzerland
t.koutsouri@sphynx.ch

Anamaria Vizitiu
*Department of Automation and Information Technology*
*Transilvania University of Brasov,* Brasov, Romania
*Advanta, Siemens SRL*
Brasov, Romania
anamaria.vizitiu@unitbv.ro

Antonios Valachis
*Department of Oncology, Faculty of Medicine and Health*
*Örebro University*
Örebro, Sweden
Antonios.Valachis@oru.se

Miriam Quintero Padron
*Atos Research and Innovation*
*Atos S.A.E*
Madrid, Spain
miriam.quintero@atos.net

*Abstract*— **The latest cancer statistics indicate a decrease in cancer-related mortality. However, due to the growing and ageing population, the absolute number of people living with cancer is set to keep increasing. This paper presents ASCAPE, an open AI infrastructure that takes advantage of the recent advances in Artificial Intelligence (AI) and Machine Learning (ML) to support cancer patients' quality of life (QoL). With ASCAPE health stakeholders (e.g. hospitals) can locally process their private medical data and then share the produced knowledge (ML models) through the open AI infrastructure.**

## I. Introduction

The latest cancer statistics [1] highlight an encouraging decrease in cancer-related mortality. Nevertheless, one in two people will be diagnosed with cancer in their lifetime. This means that, due to the growing and ageing population, the absolute number of people living with cancer will keep increasing substantially in the near future [2]. Recent scientific and technological advances give new hope to cancer patients. The ICT (Information and Communication Technologies) sector has been achieving major breakthroughs (in areas like Big Data analysis, Artificial Intelligence, Machine Learning etc.) and can now process large amounts of information to provide insights for highly complicated problems. At the same time, the digitalization of health records and the increasing use of IoT wearables, medical devices, implants etc. produce large health-related datasets that can be used to address one of the most challenging medical issue of our time.

To take full advantage of technologies like Machine Learning though, the healthcare sector still needs to address several issues, with the most important of them being the privacy of medical data. The quality of ML solutions is directly associated with the amount of high-quality data used in the training process and today, medical data still remain fragmented and isolated inside strict administrative domains. This fragmentation results in the creation of multiple -again isolated- solutions which are trained on limited and diverse datasets and are unable to take full advantage of the power of ML technologies.

This paper presents the ASCAPE framework, an open AI infrastructure designed to help healthcare providers (e.g. hospitals, private clinics etc.) co-operate and collaboratively build knowledge about cancer, without the need to share their private medical data. To achieve this, our proposed solution makes use of a) Federated Machine Learning and b) Homomorphic Encryption technologies (both analyzed in section III). Currently, ASCAPE focuses on improving the quality of life (QoL) of breast and prostate cancer patients. However, its open architecture can be used for the training of

The research leading to these results has received funding from the European Union's Horizon 2020 research and innovation programme, in the context of "Artificial intelligence Supporting CAncer Patients across Europe (ASCAPE)" Action under Grant Agreement No 875351.

any kind of ML model. This means that in the future, ASCAPE can be used to address more types of cancer or other diseases.

The rest of the paper is organized as follows. In section 2 we present the beyond the state-of-the-art aspects and in section 3 we describe the ASCAPE ecosystem. Section 4 analyses ASCAPE's technical architecture and section 5 highlights the Proof of Concept (PoC) implementation and early results on retrospective data for breast and prostate cancer. Finally, section 6 concludes this paper discussing future work.

## II. BEYOND THE STATE OF THE ART

Due to page limitation, it is not possible to extensively analyse the state of the art (SOTA) of the large number of technologies related to the ASCAPE ecosystem. Instead, in this section we present the "beyond the SOTA" aspects and main scientific advances we introduce.

Recent advances in deep learning methods based on artificial neural networks have led to breakthroughs in long-standing AI tasks such as speech, image, and text recognition. With the help of **Federated Learning**, it is possible to distribute the same ML/DL model among different actors. However, those solutions introduce privacy issues, which should be handled with care, especially when dealing with sensitive data. Proposed approaches for privacy preserving distributed learning rely on a central server and assume that the local data distribution is the same for all users. ASCAPE designed, implemented, and is evaluating a federated learning system that enables multiple parties to jointly learn an accurate machine learning model for a given objective, by applying a differentially-private scheme [3] without sharing their input datasets [4][5]. ASCAPE is based on the fact that certain optimization algorithms used in DL (e.g. the stochastic gradient-descent ones) [6], can be executed in a parallel, asynchronous, distributed manner. Also, the ASCAPE federated ML system is enhanced by differential privacy in order to guarantee user anonymity for QoL-related predictive models [7]. The developed method focuses on minimizing the information shared during the learning process, while special care is paid not only to the accuracy and speed of the method but also to the overall privacy.

To further secure its AI algorithms and the corresponding patients' data related to personal identity and health condition, ASCAPE exploits **Homomorphic Encryption**. This is a technique that allows for computations to be performed on encrypted messages without knowing the actual values hidden through the encryption. Homomorphic encryption has been recently exploited in several works for the development of AI and ML algorithms over encrypted data, such as deep neural networks (DNN). ASCAPE carries out research on how a homomorphic-encryption-based DNN model can be applied directly on floating point numbers, while incurring a reasonably small computational overhead [8]. To allow for privacy-preserving computations in the context of machine learning-based real-world medical applications, we use a simpler homomorphic encryption cryptosystem, which is based on linear transformations [9]. This class of cryptosystems, while being criticized for its security weaknesses over standard homomorphic encryption schemes, we consider to be the only viable solution for achieving real-world privacy-preserving machine learning applications.

ASCAPE also tackles one of the main challenges of AI in healthcare, and particularly for cancer. It not only provides a prediction (related to diagnosis, pathology evolution, etc.), but also estimates the uncertainty of the prediction (e.g. a correct estimation of the uncertainty may trigger the involvement of a clinical expert in case of doubt). Bayesian neural networks [10] [11] have been proposed as a solution, but it remains open how to specify their prior. Very recently, noise contrastive priors were employed successfully for obtaining reliable uncertainty estimates. Alternatively, an implicit Bayesian approximation was leveraged, that links neural networks to deep Gaussian processes, allowing for a quantification of the output uncertainty [11]. ASCAPE is pursuing this approach in an effort to develop **explainable AI solution** that allow patients and clinicians to obtain insight into how and why certain predictions are made [12]. More specifically, ASCAPE evaluates and refines feature attribution methods (e.g., SHAP [13]) to gain insight on key predictor variables and the training of surrogate models that are per se interpretable (e.g., linear regression, decision trees) to explain how predictor variables influence QoL predictions. Particular attention is given to include medical domain knowledge from experts into the process.

Finally, compared to existing healthcare projects and solutions, ASCAPE innovates in multiple levels (technical, scientific, social etc). A non-exhaustive list of projects, either completed or active, that are related to ASCAPE includes: (i) AI-MICADIS [14] which developed and tested an extremely accurate, non-invasive AI tool for early detection and diagnosis of multiple cancer types, (ii) FAITH [15], a federated AI solution for monitoring mental health status after cancer treatment, (iii) AIPACA [16] which aims to develop an AI-based software capable of analysing biopsy slides within seconds to identify tumour regions and quantify tumour biomarkers, (iv) CANCER-RADIOMICS [17] which will develop deep learning radiomic biomarkers to predict treatment response based on imaging analysis and (v) ProCAncer-I [5] which proposes to develop advanced artificial intelligence models to address unmet clinical needs: diagnosis, metastases detection and prediction of response to treatment. Further to these solutions, the clear contributions of ASCAPE framework are a) **Open ecosystem to share knowledge** (ML models) between healthcare providers without sharing actual medical data. b) **Modular agnostic architecture** that can be extended to support any type of healthcare function (diagnosis, treatment etc.) and any type of health condition. c) **Democratization of access to cancer treatment** by allowing less developed countries or small healthcare providers (small hospitals in villages) to connect to ASCAPE and benefit from its general knowledge (ML models).

## III. ASCAPE ECOSYSTEM

### A. ASCAPE approach

The ASCAPE ecosystem is formed by multiple nodes (ASCAPE edge nodes) connected to a central cloud (ASCAPE cloud). The main idea behind ASCAPE is to allow healthcare providers (e.g. hospitals, private clinics etc.) host a local node (ASCAPE Edge Node) and access "knowledge" about cancer that has being built over time by the contributions of all the nodes (other healthcare providers) that participate in ASCAPE. This "knowledge" is built via one of the following two options:

*1) Federated ML process:* With the Federated ML option, each healthcare provider installs a local node and uses its medical data to **locally** train ASCAPE's ML models (one or many). Then with the use of Federated ML technologies, the ASCAPE cloud collects all the locally trained ML models and combines them together into a "general knowledge" (Federated ML models). The Federated models are then sent back to all ASCAPE edge nodes a) as the basis for new training rounds with new data and patients b) for providing AI-assisted medical services to their patients. Federated learning can be considered as a secure-by-design, privacy-preserving machine learning technique since the knowledge is shared among everyone while the medical data remain private.

*2) Homomorphic Encryption ML process:* The training of ML models locally requires to execute complex and compute-intensive AI algorithms. For healthcare providers which cannot operate a full resource-demanding edge node, ASCAPE provides the option to securely process medical data in the cloud. For this option, ASCAPE uses state-of-the art homomorphic encryption (HE) solutions based on which the edge nodes' medical data are first being encrypted through personalized keys (keys are kept locally at the hospital). The encrypted data are passed to the cloud, which in turn performs a global model learning or inference based on the mechanisms of homomorphic encryption. Once an encrypted prediction by the HE-based model is obtained, the cloud forwards the prediction to the edge node that initiated the prediction request. The edge node subsequently decrypts the prediction that could be then combined with predictions made by the corresponding federated and local models by taking into account estimated accuracies of all those models.

*B. ASCAPE services for improving QoL of cancer patients*

Two types of services are currently offered by ASCAPE. These services are provided through a web interface, the ASCAPE Dashboard (described in section IV).

*1) General predictions:* This service constantly analyzes large numbers of diverse datasets including non-medical ones (environmental data, cost of living, demographic data, etc.) to build knowledge on how related predictions about the QoL of cancer patients are affected in relation to specific health determinants (e.g. environmental conditions, weight, gender).

*2) Patient-centric (personalized) support:* This service is designed to use as input the medical record of one specific patient and propose personalized support that will improve his/her QoL during the treatement process. The outcome is recommendations for focused interventions, as early as possible, to prevent or quickly address any kind of deviation from the desired course.

Healthcare providers are not required to contribute data to the ASCAPE ecosystem in order to gain access to its services (democratization of access to cancer knowledge). This means that small healthcare providers (e.g., in remote villages, small islands etc.) can deploy a lightweight node, download the latest ML models and take advantage of updated general predictions and recommendations for their patients without the need to maintain complex IT systems or execute AI algorithms.

*C. ASCAPE actors*

The main actors (users) of ASCAPE framework are:

*1) IT professionals:* Administrators, ICT security and AI professionals etc. These are the users assigned with the management, maintenance and technical improvements of the framework.

*2) Doctors:* Doctors will be **the actual users of the ASCAPE services** (**not patients**). Doctors are expected to always evaluate ASCAPE suggestions and decide whether they will adopt or ignore them. ASCAPE monitors which interventions are chosen and use this information to improve its efficiency. We should mention that doctors are only allowed to use ASCAPE services and predictions and cannot suggest or specify their own models in terms of predictors and target variables. The addition of new models to ASCAPE is a complex task and can only take place through a centrally coordinated channel. The process of adding new models to the ASCAPE framework is outside the scope of this paper.

*3) Patients and family:* Even though patients and their family will not directly use the ASCAPE, they will be the main beneficiaries, receiving personalized treatment and intelligent interventions (from their doctors) to improve their QoL.

*D. Explainable AI - Surrogate models*

Most Deep Learning based machine learning models that provide accurate predictions are very complex, have a large parameter space and are not interpretable by design. As explainability is an important feature of a system supporting doctors in the treatment of patients, ASCAPE adds explanations to predictive results for the models obtained through federated learning or on HE encrypted data, using feature attribution computations and explainable surrogate model training and inference. Feature Attribution for a model describes the amount each input feature contributes to the prediction of the model. A surrogate model explains the overall inner workings of a model but is interpretable by design, so it can be examined to provide explanations on the relation between input and output variables.

*E. ASCAPE data aspects*

ASCAPE collects and manages various types of data.

*1) Medical data – Retrospective:* ASCAPE uses retrospective data from epidemiological databases from specific regions of Sweden, hospital-based databases from various European hospitals, like University Hospital of Örebro (Department of Oncology), online companies for cancer support like CareAcross etc. ASCAPE's early results using retrospective data are presented in section V.

*2) Medical data – Prospective:* Prospective data will be collected through the active monitoring of patients recruited by healthcare providers collaborating with ASCAPE. This information will be obtained by various means like wearables and mobile applications, social activity information etc. ASCAPE has already started the process of recruiting patients and is expected to produce updated results in the next months.

*3) Open databases:* Apart from the medical databases, ASCAPE also supports data acquisition from external open databases like: healthcare Index per country/location, environmental, financial, socio-demographic data etc. Based on the privacy and security requirements, the collected open data (e.g. open environmental and socio-economical data) is combined with the patient data from healthcare providers' medical records and only processed locally inside edge-nodes.

Data management inside ASCAPE involves multiple processes like acquisition, consolidation, curation etc. To ensure proper usage and to avoid potential misuses and misinterpretations of the data, a judicious data annotation process is performed. The clinical information is obtainable in a scalable way, stored and served based on the definition of HL7 FHIR. The integration process also supports the combination of unstructured information sources (if any) with structured sources and the use of additional standards such as CEN / ISO EN13606, or prior versions of HL7, v2.x messages or xml CDA documents.

*F. Evaluation across multiple sites*

An initial proof-of-concept implementation of ASCAPE solution has already been completed and tested (see section V) using retrospective data from Örebro University Hospital. The first version of ASCAPE which will be released in the following months, will be evaluated in four different pilot sites.

*1) Spain:* Department of Medical Oncology, Hospital Clinic of Barcelona and Clínic Foundation for Biomedical Research (FCRB), Barcelona, ES.

*2) Sweden:* Department of Oncology, Faculty of Medicine and Health, Örebro University, SE.

*3) Greece:* 2nd Department of Urology, National and Kapodistrian University of Athens Faculty of Medicine Marousi, Athens, GR.

*4) United Kingdom:* CareAcross Ltd, London, UK

All sites will contribute both retrospective and prospective data from approx. 500 patients (in total) that are recruited for this purpose (recruitment process has already started). Also, during the second phase of evaluation, more healthcare providers will be invited (inside the context of an open call) to contribute data and test ASCAPE. The clinical protocol for the validation phase of ASCAPE has already been published on clinicaltrials.gov [18].

## IV. ASCAPE TECHNICAL ARCHITECTURE

The general architecture and internal components of ASCAPE architecture are presented in Figure 1.

*A. ASCAPE cloud*

The ASCAPE cloud hosts the necessary components to perform the following actions: a) orchestrate the Federated ML process b) collect the local ML models c) offer the HE functionality for secure processing of encrypted medical data for healthcare providers with limited resources and d) orchestrate the surrogate model training. Its components are:

*1) Cloud Federated Learning Coordinator:* The Federated Learning Coordinator centrally orchestrates the federated learning of predictive models. It also handles requests issued from ASCAPE Edge Nodes to start a new round of federated model training, e.g., due to new available training data at the edge nodes.

*2) AI Knowledge Manager:* The ASCAPE AI knowledge Manager is the central place, where all global AI models are stored. This includes both global federated learning models as well as the HE models. Each of these is combined with the interpretable surrogate model obtained from the surrogate model manager.

*3) HE Redacted Patient Manager:* This component stores all homomorphically encrypted redacted patient data obtained from ASCAPE Edge Nodes.

*4) HE AI Results Manager:* The HE AI Results Manager receives all inference requests on HE encrypted patient data to be computed by an HE model. The HE model can be retrieved from the ASCAPE Edge Node that submitted the request.

*5) HE AI Models Manager:* The HE AI Models Manager is the central place where all HE models are trained based on the available encrypted data. All trained HE models are forwarded to the ASCAPE AI Knowledge Manager and also to the Cloud Global Surrogate Models Manager.

*6) Cloud Global Surrogate Models Manager:* This component trains interpretable surrogate models for global models on the ASCAPE Cloud. It does it both for models obtained via federated learning as well as HE.

*B. ASCAPE Edge Node*

The ASCAPE Edge Node is deployed inside the premises of a healthcare provider. Its modular design provides high levels of flexibility, allowing diverse integration options with a Healthcare Information System (HIS). For instance, a hospital may choose to use ASCAPE's device data adaptors or create/use their own based on a clear ASCAPE-specified REST API. The Edge Node only accepts and stores medically relevant redacted patient data. The HIS is expected to pseudonymise all data before it sends them to the Edge Node. Thus, the Edge Node, even though it is located inside the healthcare provider's (i.e. hospital) premises, it never comes into contact with real patient information and identifiers like patient's name, National Security Number, Hospital Information System Patient Identifier etc. The ASCAPE Edge Node is focused only on providing AI capabilities, not on replicating the standard functionalities of a HIS and this is reflected both by its privacy by design architecture and the strict rules and policies of ASCAPE APIs and data models for the transmission and storage of patient data.

An ASCAPE Edge node hosts the following components:

*1) Edge AI API Gateway:* To facilitate the integration of a HIS with ASCAPE, the ASCAPE Framework defines a simple REST API with two methods, one for synchronizing patient data and one for obtaining AI analytics results.

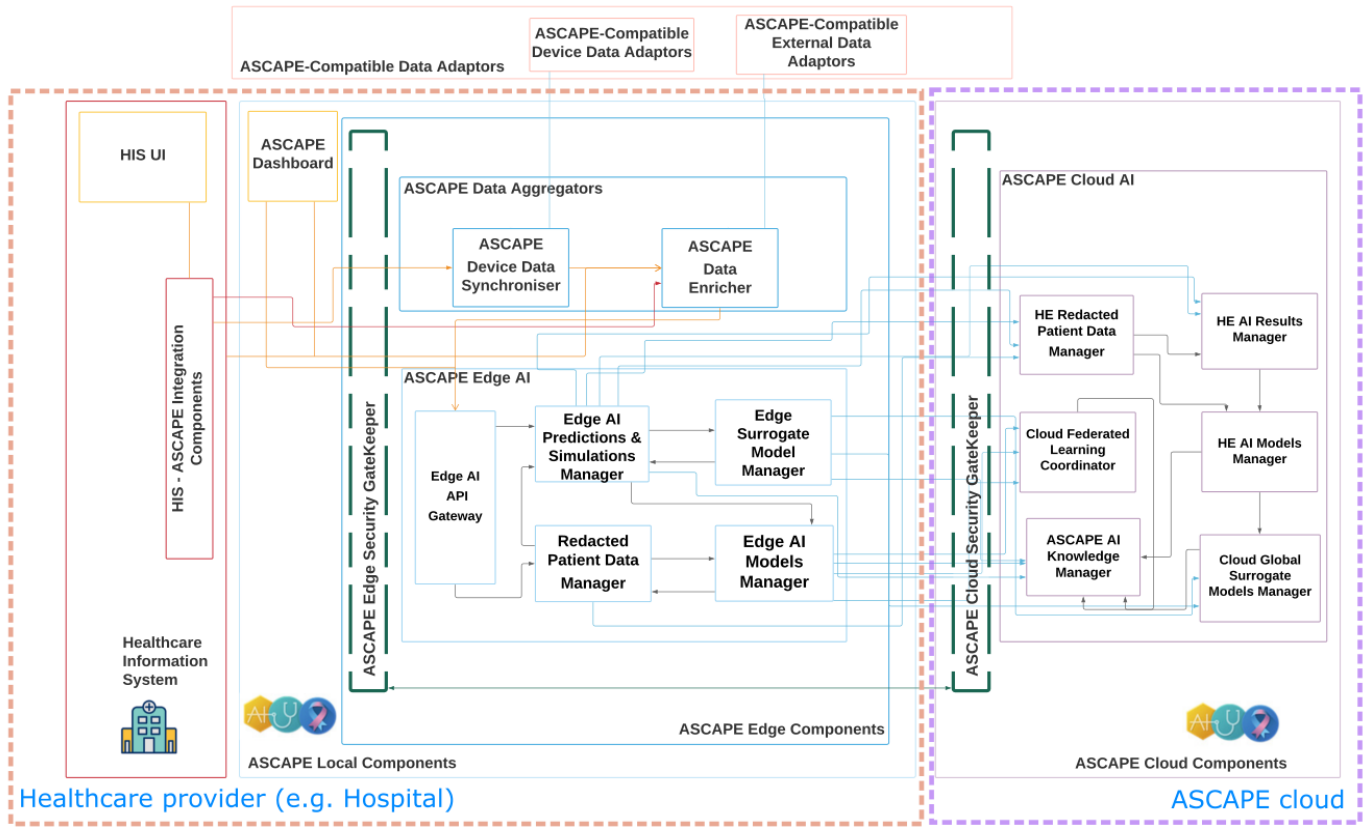

Fig. 1: ASCAPE technical architecture

*2) Redacted Patient Manager:* The Redacted Patient Data Manager is the central component receiving and processing the patient data from the Healthcare Information System and any patient devices. It also performs the following pre-processing on patient data for model training: a) performs Missing Value Inference (MVI) b) filters extreme or unlikely values, so-called outliers and c) utilises differential privacy, which introduces a small and controlled amount of noise to prevent identification of patients from trained models.

*3) Edge AI Predictions and Simulations Results Manager:* This Manager is the central component handling personalized predictions and simulations using all available AI models. It also provides, when required, explanations for these. The same component also homomorphically encrypts patient datasets and submits them (encrypted) to the ASCAPE Cloud.

*4) Edge AI Models Manager:* The Edge AI Models Manager trains the global (federated, collectively learned) and local (non-federated, non-collectively learned) QoL predictive machine learning models, as well as models for training-dataset pre-processing (AI models for missing value inference and outlier elimination). This component, together with the Cloud Federated Learning Coordinator at the cloud side, enables Federated (collective) learning of predictive QoL models.

*5) Edge Surrogate Model Manager:* This component manages the computation of interpretable surrogate models for the trained ASCAPE AI models.

*6) ASCAPE Data Enricher*: This component obtains data from external data sources (e.g. environmental and socio-demographic data) and includes them into redacted patient data.

*7) ASCAPE Device Data Synchroniser:* This component updates the redacted patient data inside the ASCAPE Edge Node with data collected by various devices (e.g. wearables, home IoT etc.).

*8) The ASCAPE Security GateKeeper:* A complete security framework is integrated into the ASCAPE Framework and is realized through Security Gatekeepers in both the Edge and Cloud sides. The main functionality provided by the Security Gatekeeper component is centralised authentication, authorisation and auditing so that all components can communicate through a secure environment.

*9) The ASCAPE Dashboard:* A web application *(Figure 2)* which doctors use to access ASCAPE functionality for a) obtaining general AI predictions and b) AI-assisted monitoring of their patients' Quality of Life (QoL) status, recording information about proposed interventions etc. The User Interface (UI) of ASCAPE Dashboard has been designed by experts in the field of healthcare IT systems with the direct involvement of clinicians and healthcare professionals. It hides the complexity of the underlying technologies and uses visualisations which are friendly to the doctors. Its offers an overview of the past, current and AI-predicted QoL issues of a patient, and provides intervention recommendations in an unobtrusive manner. Through the ASCAPE Dashboard, doctors also have the ability to obtain more details about presented visualisations and seek explanations for the AI predictions and proposed recommendations. Finally its functionality allows doctors to identify cases of patients which must be proactively contacted for initiating or modifying a specific intervention.

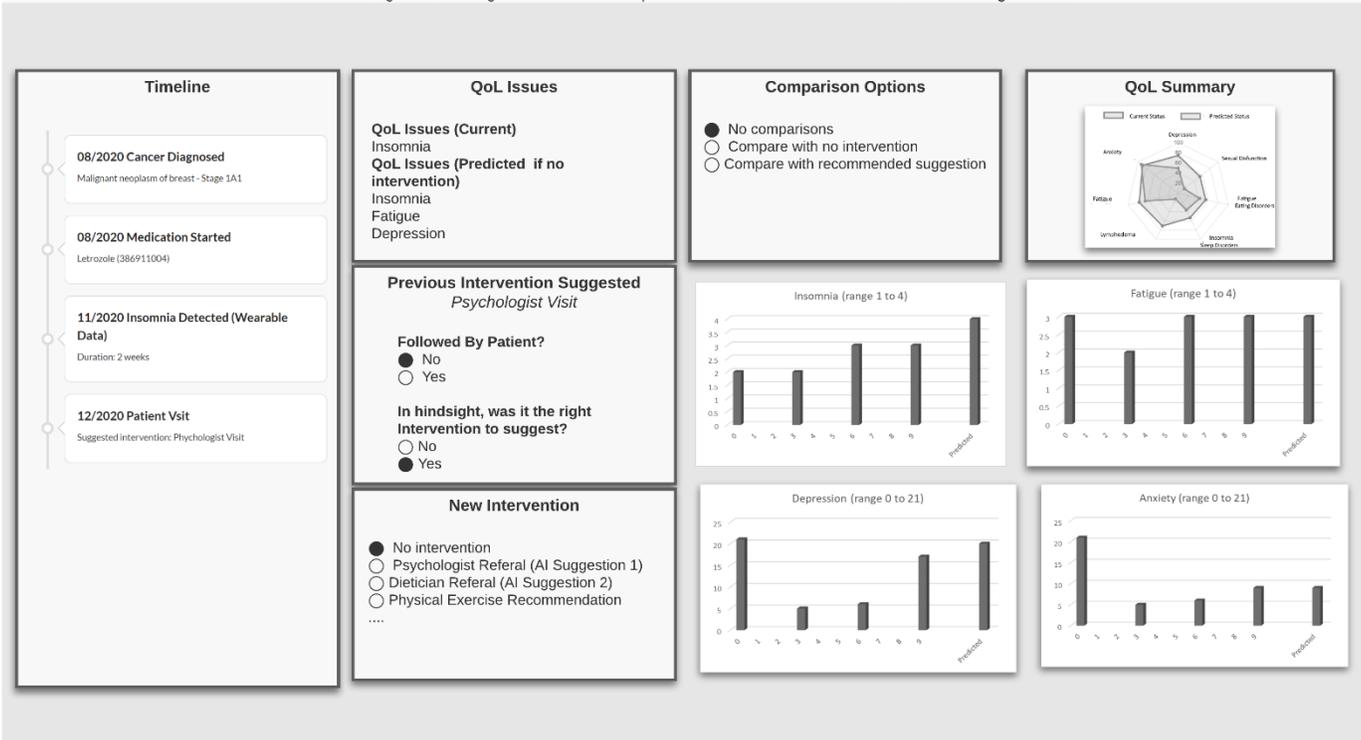

Fig. 2: ASCAPE Dashboard

## V. IMPLEMENTATION – EVALUATION

### A. Technologies used for Proof of Concept implementation

A first proof-of-concept implementation of ASCAPE has been completed and validated. For the purposes of the Proof-of-Concept Edge-Cloud Architecture, k3s [19] was used for both the Edge and the Cloud. The core ASCAPE edge node machine learning services are enabled by a set of Python modules implemented using Scikit-learn [20] and Tensorflow [21] machine learning libraries. The core ASCAPE HE-based machine learning services are enabled through an in-house developed C++ library, which is based on the MORE (Matrix Operation for Randomization or Encryption [22]) encryption implemented by Siemens [23].

### B. Datasets for training and evaluation

Our initial experiments on the proof-of-concept implementation have been based on retrospective datasets. Two datasets have been provided by Örebro University Hospital in Sweden which contained datasets with medical data of patients diagnosed with breast cancer: BcBase dataset (18988 rows x 47 columns) and prostate cancer: ORB dataset (2466 rows x 124 columns).

The BcBase dataset contains patient data of 18.988 breast cancer patients diagnosed in three different healthcare Regions in Sweden with 47 variables per sample. The dataset also contains socio-economic indicators like marital status, education status, personal income and household income. The Örebro dataset (ORB) contains 2466 health records of prostate cancer patients with each containing 124 variable fields collected at the time of diagnosis and scheduled follow-ups after 3 weeks and then again in months 6, 12, 18, 24, 30, 36, 42, 48, 54, 60, 72, 84, 96, 108 and 120. In the follow-ups the patient reported about bowel side effects, erectile function and lower urinary tract symptoms. Additionally, for the follow-up dates at months 36, 60 and 120, the International Prostate Symptom Score (IPSS) and the Lisat Quality of Life Score was collected from the patients.

For the proof-of-concept phase, the retrospective datasets were not transformed to the HL7 FHIR format but were arranged as a table saved as CSV-format. Each row contains medical data of one patient and each column contains one variable, representing a field in a patient's medical data. A variable can be a date or duration, categorical data, ordinal data, information about a patient like the age of diagnosis, scores for standardized medical questionnaires and measurements done during treatment etc. The ORB dataset contains QoL scores at the time of the diagnosis and three different times relative to the date of diagnosis at months 36, 60 and 120. The dataset was used to train models to predict future QoL scores and create a variety of datasets ORB-*n-m*. An example of such a dataset is ORB-30-

120 which is used to predict the QoL score for month 120 based on all variables that are available up until month 30. The same scheme is used to create the experiment datasets ORB-30-36, ORB-30-60, ORB-30-120, ORB-54-60, ORB-54-120 and ORB-108-120, respectively (6 databases). The BcBase dataset does not have QoL scores but contains information whether medications to treat pain, anxiety, insomnia, or depression were given to the patient. This dataset is then used to create datasets to train binary classifiers which estimate if a patient has one of these conditions. The classification scores of the trained models can be used for risk assessment. The new datasets were created by dropping all but one medication variable and keeping all other variables to obtain the datasets BcBase-Anxiety, BcBase-Pain, BcBase-Insomnia and BcBase-Depression (4 databases).

The first actions for the preparation of these 10 datasets for the machine learning algorithms were to impute empty data points and to create new datasets from applying differential privacy to ensure privacy.

*C. Model training and evaluation*

The evaluation process examined a large number of algorithms against all databases. Due to page limitations we cannot include all the details of the evaluation process and will present detailed results of only one database for breast (BcBase-Anxiety) and one for prostate cancer (ORB 30-60). For other datasets we obtained highly similar results. In the experimental evaluation we examined the most classic and widely used machine learning models for classification and regression. Each ASCAPE predictive model was internally validated; classification-based models by computing precision/recall/F1 scores, regression-based models by computing MAE/MSE/correlation between predicted and real (ground truth) values of QoL indicators. Those evaluation results provide confidence scores for predictions for each particular model.

*1) Training and evaluation of Local AI Models:* An ASCAPE edge node can use its own local models for making predictions instead of the corresponding global models when the global models exhibit a poor performance (low accuracy or high error) on training datasets present on that node. The following machine learning algorithms are considered for training local models performing classification: **SVM**: support vector machine classifier, **NB**: Naïve Bayes classifier, **KNN**: K-nearest neighbours' classifier (K=10), **DT**: Decision-tree classifier, **RF**: random forest classifier. For regression problems, local models are trained by one of the following machine learning algorithms: **LINEAR**: linear regression, **RIDGE**: ridge regression, **LASSO**: lasso regression, **ELASTICN**: elastic net regression, **KRIDGE**: kernel ridge regression, **SVR**: support vector machine regression, **RF**: random forest regression, **KNN**: K-nearest neighbours regression, **ADAB**: AdaBoost regression. As the baseline for evaluating above above-mentioned regression models we use the so-called **DUMMY** regression model. The DUMMY model always predicts the same value: the mean of the outcome variable computed from the training dataset.

For the evaluation of models' performance, we used 10-fold cross validation. In order to assess the performance of models from different aspects, various model evaluation measures are examined. For the problem of binary classification those are: accuracy ($ACC$), F1 score ($F1$), precision of the positive class ($Prec^+$), recall of the positive class ($Rec^+$), precision of the negative class ($Prec^-$), recall of the negative class ($Rec^-$), macro-averaged precision ($Prec$), and macro-averaged recall ($Rec$). All of the mentioned measures are formally defined below, where $tp$ stands for true positives, $tn$ for true negatives, $fp$ for false positives, and $fn$ for false negatives.

$$ACC = \frac{tp + tn}{tp + tn + fp + fn}, \qquad F1 = \frac{2tp}{2tp + fp + fn}$$

$$Prec^+ = \frac{tp}{tp + fp}, \qquad Rec^+ = \frac{tp}{tp + fn}$$

$$Prec^- = \frac{tn}{tn + fn}, \qquad Rec^- = \frac{tn}{tn + fp}$$

$$Prec = \frac{Prec^+ + Prec^-}{2}, \qquad Rec = \frac{Rec^+ + Rec^-}{2}$$

On the other side, the evaluation of the regression models is based on mean absolute error ($MAE$), mean squared error ($MSE$), coefficient of determination ($R2$), and the Person's correlation coefficient ($PC$). All the measures are presented below, where $n$ is the number of dataset instances, $x_i$ is the target attribute value of the $i$-th instance, $y_i$ is the predicted value of the target attribute of the $i$-th instance, $\bar{x}$ is the mean of the target attribute's values, and $\bar{y}$ is the mean value of all predicted values for the target attribute.

$$MAE = \frac{\sum_{i=1}^{n}|x_i - y_i|}{n}$$

$$MSE = \frac{\sum_{i=1}^{n}(x_i - y_i)^2}{n}$$

$$SS_{tot} = \sum_{i=1}^{n}(x_i - \bar{x})^2, \qquad SS_{res} = \sum_{i=1}^{n}(x_i - y_i)^2,$$

$$R2 = 1 - \frac{SS_{res}}{SS_{tot}}$$

$$PC = \frac{\sum_{i=1}^{n}(x_i - \bar{x})(y_i - \bar{y})}{\sqrt{\sum_{i=1}^{n}(x_i - \bar{x})^2}\sqrt{\sum_{i=1}^{n}(y_i - \bar{y})^2}}$$

The evaluation of centrally-trained binary classification models on the BcBase datasets indicated that NB is the best performing centrally trained model. Table I presents the evaluation results for binary classification models on BcBase-Anxiety.

TABLE I. EVALUATION RESULTS FOR BINARY CLASSIFICATION MODELS ON BCBASE-ANXIETY

|     | ACC | F1 | Prec | Rec | $Prec^+$ | $Rec^+$ | $Prec^-$ | $Rec^-$ |
|-----|-----|-----|------|-----|-------|------|-------|------|
| RF  | 0.673 | 0.484 | 0.535 | 0.514 | 0.366 | 0.112 | 0.704 | 0.916 |
| SVM | 0.698 | 0.411 | 0.349 | 0.500 | 0.000 | 0.000 | 0.698 | 1.000 |
| **NB** | **0.629** | **0.552** | **0.553** | **0.552** | **0.379** | **0.354** | **0.728** | **0.749** |
| KNN | 0.682 | 0.458 | 0.529 | 0.507 | 0.357 | 0.066 | 0.701 | 0.949 |
| DT  | 0.583 | 0.511 | 0.511 | 0.511 | 0.317 | 0.329 | 0.705 | 0.693 |

It can be seen that SVM is the model with the highest accuracy. However, this model achieves both zero precision and recall for the positive class (patients experiencing negative QoL-related symptoms). In other words, SVM is unable to provide accurate predictions for the positive class and its high accuracy is actually the consequence of class-imbalanced training datasets (approximately 70% of data instances belong to the negative class and 30% to the positive class). Thus, it is much better to use $F1$ score (metric that aggregates precision and recall scores of both classes into a single score) to compare different models. The largest $F1$ score is achieved by NB since it has the highest precision and recall scores for the positive class. For all classification models we can see relatively low level of precision and recall for the positive class. Thus, in our future work we will examine adequate random sampling techniques to create more class-balanced samples for training the examined models in order to improve their performance for the positive class, and consequently increase their overall performance (in terms of $F1$ scores).

The evaluation of centrally-trained regression models on the ORB datasets revealed that the best performing model is LASSO. Table II presents the evaluation results for regression models on ORB-30-60.

TABLE II. EVALUATION RESULTS FOR REGRESSION MODELS ORB-30-60

|  | MAE | MSE | R2 | PC |
|---|---|---|---|---|
| DUMMY | 6.890 | 78.402 | -0.001 | NaN |
| LINEAR | 6.129 | 71.126 | 0.086 | 0.429 |
| RIDGE | 5.925 | 66.454 | 0.147 | 0.464 |
| **LASSO** | **5.886** | **62.182** | **0.205** | **0.463** |
| ELASTICN | 5.913 | 62.530 | 0.201 | 0.460 |
| KRIDGE | 5.958 | 67.246 | 0.137 | 0.459 |
| SVR | 6.773 | 80.572 | -0.028 | 0.039 |
| RF | 6.015 | 62.576 | 0.202 | 0.459 |
| KNN | 6.968 | 80.045 | -0.023 | 0.076 |
| ADAB | 6.542 | 67.651 | 0.135 | 0.436 |

*2) Training and evaluation of simulated federated models:* ASCAPE federated learning services were experimentally evaluated on simulated ASCAPE edge nodes collectively producing federated models either in the incremental or semi-concurrent federated learning mode. In the incremental mode, federated model updates are performed sequentially from the first to the last ASCAPE edge node. In the semi-concurrent mode, ASCAPE edge nodes simultaneously update the model producing different instances of the model that are averaged for the next round of updates. For each experimental dataset we trained and evaluated several federated models for both federated learning modes and for different number of simulated edge nodes (from 2 to 4) by the following procedure.

First, the dataset was divided into 10 folds for the purpose of the 10-fold cross-validation. Then, the following steps were performed in each iteration of the 10-fold cross-validation:

*a)* The training part of the dataset (9 folds) was randomly split into k equally-sized stratified splits, where k is the number of simulated ASCAPE edge nodes. The obtained splits were assigned to simulated ASCAPE edge nodes as their local datasets.

*b)* The simulated federated model was collectively trained by simulated ASCAPE edge nodes each of them using its own split to perform model updates.

*c)* The trained model was evaluated on the test fold by computing appropriate evaluation metrics (one set of metrics for regression-based models and the other set of metrics for classification-based models as described in the subsection "*Training and evaluation of Local AI Models*" above).

In experiments with simulated federated models, we have used different neural network architectures for different datasets. A preliminary investigation, in which we have varied the number of hidden neural network layers between 1 and 10 and the batch size in the set {16, 64, 128, 256, 512}, showed that shallow neural networks (a small number of hidden layers) trained with a large batch size are more suitable for the BcBase datasets, while deeper neural networks (a larger number of hidden layers) trained with a small batch size result with better predictive models for the ORB datasets. We have simulated from 2 to 4 ASCAPE edge nodes training models in both incremental and semi-concurrent federated learning mode. The comparison of $F1$ scores of local and simulated federated binary classification models on the BcBase datasets is presented in Table III. TFNN denotes a local TensorFlow-based neural network binary classification model, while INC-k and CON-k are simulated federated TensorFlow-based neural network binary classification models trained in the incremental (INC) and semi-concurrent (CON) learning mode for k simulated edge nodes.

For the BcBase-Anxiety, Depression and Insomnia datasets, we have identified that simulated federated models are significantly better than the worst performing local model (SVM and KNN depending on the dataset). The $F1$ scores of simulated federated models are close to $F1$ scores of NB which is the best performing local model for those three BcBase datasets.

TABLE III. COMPARISON OF F1 SCORES OF LOCAL AND SIMULATED FEDERATED BINARY CLASSIFICATION MODELS ON THE BCBASE DATASETS

|  | Anxiety | Depression | Insomnia | Pain |
|---|---|---|---|---|
| Best local | 0.552 | 0.534 | 0.554 | 0.522 |
| Worst local | 0.411 | 0.413 | 0.502 | 0.457 |
| TFNN | 0.438 | 0.530 | 0.540 | 0.542 |
| INC-2 | 0.536 | 0.512 | 0.546 | 0.542 |
| INC-3 | 0.542 | 0.507 | 0.529 | 0.542 |
| INC-4 | 0.539 | 0.515 | 0.538 | 0.532 |
| CON-2 | 0.522 | 0.504 | 0.542 | 0.548 |
| CON-3 | 0.512 | 0.519 | 0.550 | 0.534 |
| CON-4 | 0.530 | 0.509 | 0.542 | 0.545 |

For simulated federated regression models trained on the ORB dataset (Table IV), we have used the neural network architecture with 10 hidden layers each with 40 neurons. Their training was performed in 200 epochs per simulated ASCAPE edge node. The batch size was equal to 32. The optimization algorithm was Adam with the same learning rate as for classification models.

TABLE IV. COMPARISON OF MAE SCORES OF LOCAL AND SIMULATED FEDERATED REGRESSION MODELS ON THE ORB DATASETS

|  | 30-36 | 30-60 | 30-120 | 54-60 | 54-120 | 108-120 |
|---|---|---|---|---|---|---|
| DUMMY | 6.541 | 6.890 | 6.909 | 6.890 | 6.909 | 6.909 |
| LASSO | 5.089 | 5.886 | 6.478 | 4.840 | 6.180 | 5.437 |
| TFNN | 5.783 | 6.572 | 7.323 | 5.811 | 7.206 | 6.562 |
| INC-2 | 6.012 | 6.775 | 7.488 | 5.931 | 7.188 | 6.625 |
| INC-3 | 6.472 | 6.751 | 7.226 | 5.867 | 7.169 | 6.430 |
| INC-4 | 6.595 | 7.042 | 7.463 | 6.220 | 7.206 | 6.484 |
| CON-2 | 5.881 | 6.705 | 7.444 | 5.904 | 7.193 | 6.463 |
| CON-3 | 6.404 | 6.826 | 7.427 | 5.986 | 7.098 | 6.327 |
| CON-4 | 6.534 | 6.883 | 7.538 | 6.269 | 7.221 | 6.652 |

For all six datasets, the best local model (LASSO) has slightly lower prediction errors than simulated federated models. There are no large differences between simulated federated models trained in different federated learning modes.

*3) Training and evaluation of HE models:* All our experiments on HE neural network models rely on multilayer perceptron (MLP). As MLPs are universal function approximators they are suitable for both classification and regression problems where a class or a real-valued quantity is predicted given a set of inputs. They are typically comprised of one or more layers of neurons. To learn a non-linear mapping from inputs to outputs, neurons make use of non-linear functions. Moreover, depending upon the activation function of the neurons in the output layer, the model is trained either to perform classification or regression.

For the BcBase datasets, the topology of the MLP classifiers consists of five layers of neurons: an input layer, three hidden layers, and an output layer. Rectifier linear unit (ReLU) is used as a non-linear activation function in the hidden layers and sigmoid function for the output. The models use the following hyperparameters for training: a learning rate of 0.01, a batch size of 128, around 300 epochs, and Stochastic Gradient Descent (SGD) for the optimizing algorithm. The objective function minimized during training is the categorical cross-entropy.

For the ORB datasets, the topology of the MLP regressors consists of seven layers of neurons: an input layer, five hidden layers with 100 neurons each, and an output layer. Hyperbolic tangent is used as a non-linear activation function in the hidden layers and a linear function for the output. The models use the following hyperparameters for training: a learning rate of 0.01, a batch size of 128, around 100 epochs, and Stochastic Gradient Descent (SGD) for the optimizing algorithm. The objective function minimized during training is the Mean Squared Error (MSE).

For the evaluation of HE-based model, all neural network models were trained on plaintext data as well. For consistency, and for enabling a fair comparison, the same architectures, hyperparameters, and random initializations were adopted. The main objective of the evaluation is to assess the practical feasibility of HE and to determine the performance impact of the use of the HE technique in machine learning-based analysis when compared to plaintext analysis. Hence, the outputs of the plaintext models are compared to the results of the encrypted models after decryption.

The performances obtained by the models trained on plaintext and ciphertext data for the classification problem on the BcBase Anxiety dataset are depicted in Table V and for the regression problem on the ORB-30-60 dataset in Table VI.

TABLE V. EVALUATION OF BINARY CLASSIFICATION MODELS ON BCBASE-ANXIETY.

|  | ACC | F1 | Prec | Rec | Prec⁺ | Rec⁺ | Prec⁻ | Rec⁻ |
|---|---|---|---|---|---|---|---|---|
| **Plaintext** | 0.653 | 0.223 | 0.514 | 0.525 | 0.865 | 0.705 | 0.164 | 0.346 |
| **Encrypted** | 0.653 | 0.223 | 0.514 | 0.525 | 0.865 | 0.705 | 0.164 | 0.346 |

TABLE VI. EVALUATION OF REGRESSION MODELS ON ORB-30-60.

|  | MAE | MSE | R2 | PC |
|---|---|---|---|---|
| **Plaintext** | 7.500 | 93.420 | -0.191 | 0.170 |
| **Encrypted** | 7.500 | 93.420 | -0.191 | 0.170 |

Due to the MORE scheme's homomorphic properties, its direct applicability on floating-point data, and its noise-free nature, an unlimited number of operations can be performed on ciphertext data without losing precision. As a consequence, operations performed on plaintext and ciphertext produce identical results (after decryption). This could be seen in our experiments, where model training progresses similarly on both plaintext and ciphertext. Hence, models trained and evaluated on MORE homomorphically encrypted data are statistically not discernible from that obtained by the models on unencrypted data. Compared to the TensorFlow models the HE-based models obtained weaker results. This is a consequence of the fact that the C++ HE AI library can operate, for the moment, only with the SGD optimizer. The performance of any predictive model generally improves with more data and favorable and optimization algorithms. The suitability of the MORE encryption scheme was also investigated in real machine learning applications, and, on average, the encrypted model took 40 times longer to train and 35 times longer to test than the unencrypted model (Table VII).

TABLE VII. TRAINING AND PREDICTION TIME FOR THE UNENCRYPTED AND THE ENCRYPTED MODEL.

| Dataset | Training time on plaintext data (s) | Training time on encrypted data (s) | Testing time on plaintext data (s) | Testing time on encrypted data (s) |
|---|---|---|---|---|
| **BcBase-Anxiety** | 448.803 | 18525.69 | 0.049 | 1.764 |
| **ORB-30–36** | 67.638 | 2675.353 | 0.018 | 0.622 |

## VI. CONCLUSIONS AND FUTURE WORK

In this paper we described ASCAPE, an open AI framework designed to enable healthcare providers (e.g. hospitals, private clinics etc.) co-operate and collaboratively build knowledge about cancer, without sharing their private data. We described its technical architecture, PoC implementation and presented initial results based on retrospective data. Currently ASCAPE is finalizing the first version of the framework architecture and has already started patient recruitment and collection of prospective data. Future work will focus on the deployment of ASCAPE solution inside healthcare providers (hospitals) to be tested by doctors. This will allow us to perform an extensive evaluation over ASCAPE's accuracy, efficiency, acceptability performance etc.